\documentclass{article}

\PassOptionsToPackage{numbers}{natbib}

\usepackage{neurips_2022}

\usepackage[utf8]{inputenc} 
\usepackage[T1]{fontenc}    
\usepackage{hyperref}       
\usepackage{url}            
\usepackage{booktabs}       
\usepackage{amsfonts}       
\usepackage{nicefrac}       
\usepackage{microtype}      
\usepackage{xcolor}         
\usepackage{todonotes}
\usepackage{array}
\usepackage{dsfont}
\usepackage[title]{appendix}

\usepackage{graphicx} 
\usepackage{algorithm}
\usepackage{algpseudocode}
\usepackage{amsmath,amssymb}
\usepackage{amsthm}
\usepackage[font=small,skip=0pt]{caption}
\theoremstyle{definition}

\title{Policy Optimization with Advantage Regularization for Long-Term Fairness in Decision Systems}
\author{
  Eric Yang Yu \\
  UC San Diego \\
  \And
  Zhizhen Qin \\
  UC San Diego \\
  \And
  Min Kyung Lee \\
  UT Austin \\
  \And
  Sicun Gao \\
  UC San Diego \\
}

\begin{document}

\maketitle

\begin{abstract}
Long-term fairness is an important factor of consideration in designing and deploying learning-based decision systems in high-stake decision-making contexts. Recent work has proposed the use of Markov Decision Processes (MDPs) to formulate decision-making with long-term fairness requirements in dynamically changing environments, and demonstrated major challenges in directly deploying heuristic and rule-based policies that worked well in static environments. We show that policy optimization methods from deep reinforcement learning can be used to find strictly better decision policies that can often achieve both higher overall utility and less violation of the fairness requirements, compared to previously-known strategies. In particular, we propose new methods for imposing fairness requirements in policy optimization by regularizing the advantage evaluation of different actions. Our proposed methods make it easy to impose fairness constraints without reward engineering or sacrificing training efficiency. We perform detailed analyses in three established case studies, including attention allocation in incident monitoring, bank loan approval, and vaccine distribution in population networks. 
\end{abstract}

\section{Introduction}
Learning-based algorithmic decision systems are increasingly used in high-stake decision-making contexts. A critical factor of consideration in their design and deployment is to ensure fairness and avoid disparate impacts on the marginalized populations~\cite{DBLP:journals/csur/MehrabiMSLG21}. Although many approaches have been developed to study and ensure fairness in algorithmic decision systems ~\cite{DBLP:journals/csur/MehrabiMSLG21, corbett2018measure}, most of the literature studies fair decision-making in a one-shot context, meaning they make the decision that maximizes fairness in a static setting. This approach fails to explicitly address how decisions made in the present may affect the future status and behaviors of targeted groups, which in turn can form a feedback loop that negatively impacts the effectiveness and fairness of the decision-making strategies. In other words, the implications associated with \emph{long-term fairness}, or fairness evaluated over a time horizon rather than in a single time step, are largely under-studied.

The long-term impact of such decision systems has recently been explored through explicit modeling of the dynamics and feedback effects in the interactions between the decision-makers and the targeted populations~\cite{DBLP:conf/fat/DAmourSABSH20,nokhiz2021precarity, zhang2020qualification, mansoury2021graphrec, creager2020causal}. In particular, the recent work of~\cite{DBLP:conf/fat/DAmourSABSH20} has demonstrated, with concrete simulation examples, how long-term fairness can not be analyzed in closed forms, but requires the use of more computational analysis tools based on simulations. They proposed to formulate such long-term dynamics and the interaction between the decision-making and the environment in the framework of Markov Decision Processes (MDPs). This formulation and the corresponding simulation environments make it possible to take advantage of recent advances in deep reinforcement learning (RL) for finding new decision-making policies that can achieve both better overall utility and fairness, compared to manually designed heuristic and rule-based strategies.

One challenge of directly using RL-based methods for learning decision-making policies, however, is that the goal of decision systems in the high-stake decision-making context is often inherently multi-objective. On one hand, from a utilitarian perspective, an effective policy should try to maximize the overall expected utility of the decisions for all targeted groups. On the other hand, constraints such as the fairness requirements should be explicitly enforced to prevent biased policies that negatively treat certain groups in temporary or historically disadvantageous situations. Since the standard RL framework for policy optimization only optimizes a policy with respect to a monolithic reward function, it can be difficult to enforce these fairness requirements during training. An intuitive approach for enforcing fairness in standard RL is to define a penalty term in the objective function that captures the magnitude of violation of the fairness requirements by the policy. However, this approach would require the RL user to define the trade-off between the utilitarian objective and the fairness objective, typically as weights on each objective. This requirement of \emph{reward engineering} can make it hard to justify the policies obtained by the RL algorithms, because one can question whether the predefined weights have introduced problematic assumptions and trade-offs between the objectives in the first place. The monolithic reward definitions may also incentivize the learning agent to perform \emph{reward hacking}~\cite{pan2022the}, or adopt undesirable behaviors that exploit the wrong trade-off between the different objectives, such as conservatively accumulating incremental rewards without achieving the overall goal just to avoid the penalty of violating constraints.

One approach for addressing such problems is the framework of Constrained Markov Decision Processes (CMDPs)~\cite{altman1999constrained}, which allows the RL user to explicitly declare rewards and constraints and use RL algorithms such as Constrained Policy Optimization (CPO)~\cite{achiam2017constrained} to simultaneously maximize reward and minimize constraint violation over time. However, the CMDP formulation only requires the learning algorithms to lower the expectation of constraint violation asymptotically, i.e. achieving constraint-abiding policies in a probabilistic sense if training time is allowed to be infinitely long, and can not ensure fairness for policies trained in practice. Moreover, in comparison to standard policy optimization methods such as Proximal Policy Optimization (PPO)~\cite{ppo}, algorithms for CMDPs can take significantly longer, and the policies obtained after finite training periods may still have high constraint-violation rates and poor performance.

We propose new methods for enforcing long-term fairness properties in decision systems by taking a constrained RL approach. At a high level, we enforce fairness requirements at the policy gradient level during policy optimization with minimal additional computational overhead. 
By enforcing fairness constraints through advantage regularization rather than at the objective level, we avoid reward engineering or hacking on the decision problems and aim to algorithmically optimize the trade-off between utility and fairness. At the same time, the proposed learning algorithms can train the decision policies much more efficiently than existing CMDP methods. Finally, the simplicity of our approach enables easy integration with off-the-shelf policy optimization algorithms in RL. Our methods are inspired by Lyapunov stability methods for improving stability in control systems~\cite{lyapunovbook, Richards2018TheLN, LIU2020108758, bobiti2018stabilityverification, DBLP:conf/nips/ChangRG19,DBLP:conf/icra/ChangG21}. We show that fairness properties can be handled in a similar framework with our specific design of the constraint regularization terms. In sum, our main contributions are as follows:

\noindent$\bullet$ We show that RL approaches are effective for designing policies that can achieve long-term fairness, where existing heuristic and rule-based approaches do not perform well~\cite{DBLP:conf/fat/DAmourSABSH20}. Specifically, we demonstrate that policy optimization methods can find strictly better decision policies that achieve higher overall utility and violate less of the fairness requirements than previously-known strategies. 

\noindent$\bullet$ We propose novel methods for imposing fairness requirements in standard policy optimization procedures by regularizing the advantage evaluation during the policy gradient steps. This approach uses control-theoretic frameworks to enforce fairness constraints and avoids reward engineering issues in the decision-making context~\cite{DBLP:conf/fat/DAmourSABSH20}.

We evaluate our approaches in several established case studies using the simulation environments~\cite{DBLP:conf/fat/DAmourSABSH20, DBLP:journals/corr/atwood2019treatment}, such as incident monitoring, bank loan approval, and vaccine distribution for infectious diseases in population networks. We find that the proposed policy optimization with advantage regularization is able to find policies that perform better than previously-known strategies, both in achieving higher overall utility and lower violation of the fairness requirements in all the case study environments.

\section{Related Work}

\paragraph{Long-term Fairness in Algorithmic Decision-Making.} The work in~\cite{DBLP:conf/fat/DAmourSABSH20} is the first to formulate long-term fairness problems in decision systems as Markov Decision Processes (MDPs). The simulation environments proposed in the work allow us to consider the agent design problem in ways that are to other RL problems such as robot control. Others have also shown that long-term fairness is nontrivial, and analyzing it in the context of a static scenario can be harmful because it contradicts fairness objectives optimized in static settings \cite{hu2019manipulation, liu2019delay, milli2019social}. For example, \cite{hu2019manipulation, milli2019social} find that providing a direct subsidy for a disadvantaged group with the purpose of improving some institutional utility actually widens the gap between advantaged and disadvantaged groups over time, which further shows that long-term fairness is difficult to achieve. There have been a growing number of studies on fairness in the long-term with various algorithmic approaches~\cite{nokhiz2021precarity, zhang2020qualification, mansoury2021graphrec, creager2020causal}. \cite{mansoury2021graphrec} proposes a graph-based algorithm to improve fairness in recommendations for items and suppliers. They relate fairness to breaking the perpetuation of bias in the interactions between users and items. \cite{creager2020causal} proposes the use of causal directed acyclic graphs (DAGs) as a paradigm for studying fairness in dynamical systems. They argue that causal reasoning can help improve the fairness of off-policy learning, and if the underlying environment dynamics are known, causal DAGs can be used as simulators for the models in training. \cite{zhang2020qualification} provides a framework for studying long-term fairness and finds that static fairness constraints can either promote fairness or increase disparity between advantaged and disadvantaged groups in dynamical systems. \cite{geandliu2021recommendation} studies how to maintain long-term fairness on item exposure for the task of splitting items into groups by recommendation, using a modified Constrained Policy Optimization (CPO) procedure~\cite{achiam2017constrained}. \cite{chi2022parity} introduces the fairness notion of return parity, a measure of the similarity in expected returns across different demographic groups, and provides an algorithm for minimizing this disparity.

Several recent works have also considered fairness-like constraints in deep reinforcement learning in various different contexts. \cite{chen2021acrl} designs fairness optimized actor-critic algorithms in deep reinforcement learning. They enforce fairness by multiplicatively adjusting the reward for fairness utility optimization in standard actor-critic reinforcement learning. The work of \cite{siddique2020momdp} studied multi-dimensional reward functions for MDPs motivated by fairness and equality constraints, and performed theoretical analysis on the approximation error with respect to the optimal average reward. Our focus is on proposing a practical algorithm for making fair decisions in the dynamic environments formulated in~\cite{DBLP:conf/fat/DAmourSABSH20} and show that policy optimization through advantage regularization can find the neural network policies that significantly outperform previously known strategies in the dynamic setting. 

\paragraph{Policy Optimization under Constraints.} The most widely-adopted formulation of RL with a set of constraints is constrained Markov Decision Processes (CMDPs) \cite{altman1999constrained, yu2019convergent}. Safety constraints are incorporated by augmenting the standard MDP framework with constraints over expectations of auxiliary costs. When models are known in discrete tabular settings, a CMDP is solvable using linear programming (LP) \cite{altman1999constrained}. However, results are limited for model-free scenarios where model dynamics are unknown, and for large-scale or even continuous state action spaces \cite{achiam2017constrained, chow2018lyapunov, yu2019convergent}. More importantly, both objective and constraint in high-dimensional CMDP settings, where high-capacity function approximators are adopted, are non-convex. Recent methods in solving CMDPs in continuous spaces can be divided into two categories, in terms of ways to incorporate constraints. In soft constrained RL, it is a common practice to apply Lagrangian method with learnable Lagrangian multipliers and solve the converted unconstrained saddle-point optimization problem using policy-based methods \cite{bohez2019value, chow2017risk, tessler2018reward}. Such Lagrangian methods achieve overall safety when policies converge asymptotically, nevertheless allowing possible violations during training. On the contrary, hard-constrained RL aims to learn safe policies throughout training. Representative works include Constrained Policy Optimization (CPO) based on trust region \cite {achiam2017constrained}, surrogate algorithms with stepwise \cite{dalal2018safe} and super-martingale \cite{mossalam2016multi} surrogate constraints, as well as Lyapunov-based approaches \cite{chow2018lyapunov, chow2019lyapunov}. 

\section{Policy Optimization with Advantage Regularization}

In long-term fairness studies \cite{DBLP:conf/fat/DAmourSABSH20}, fairness is evaluated over a time horizon where the agent interacts with the environment, and the environment can change in response to the interactions. Simulations following the MDP framework is one way to analyze fairness over time and systematically come up with strategies for maximizing fairness in the long-term rather than in a single step. MDPs naturally incorporate the idea that actions made in the present can have accumulating consequences on the environment over time. Long-term fairness is evaluated with metrics that describe the consequences made by an agent's policy on the different subgroups in an environment over time. These metrics are computed at each step of the MDP, and include data collected from the past time steps.

\paragraph{Policy Optimization in MDPs.} Formally, an MDP is defined as $M =\langle\mathcal{S}, \mathcal{A}, f, r, \gamma\rangle$ with the following components. $\mathcal{S}$ denotes the state space, and $\mathcal{A}$ the action space. 
The transition function $f: \mathcal{S} \times \mathcal{S} \times \mathcal{A} \rightarrow [0,1]$ determines the probability $f(s' |s,a) $ of transitioning into state $s'$ from state $s$ after taking action $a$. We consider general forms of reward functions 
$r: \mathcal{S}\times \mathcal{A}\times \mathcal{S} \rightarrow \mathbb{R}$ defined over transitions. We write $\pi_\theta$ to denote a stochastic policy $\pi_\theta: \mathcal{S} \times \mathcal{A} \rightarrow [0,1]$ parameterized by $\theta$. The goal of policy optimization is to maximize the expected $\gamma$-discounted cumulative return 

\begin{equation}
J(\theta) = \mathbb{E}_{s_0, a_0, \ldots}\left[\sum_{t=0}^{\infty} \gamma^t r(s_t, a_t, s_{t+1})\right].
\end{equation}

Policy optimization methods estimate the policy gradient and use stochastic gradient ascent to directly improve policy performance. A standard gradient estimator is to use $\nabla_\theta \log \pi_\theta(a_t|s_t) \hat{A}_t$, where $\pi_\theta$ is a stochastic policy and $\hat{A}_t$  estimates the advantage that represents the difference between the Q value of an action compared with the expected value of a state, to indicate whether an action should be taken more frequently in the future. The gradient steps will move the distribution over actions in the right direction accordingly. The expectation is estimated by the empirical average over finite batch of samples. The proximal policy optimization algorithm (PPO) \cite{ppo} applies clipping to the objective function to remove incentives for the policy to change dramatically, using:

\begin{equation}
J^{\rm CLIP}(\theta) \gets \hat{\mathbb{E}}\left[\min\left(R_t(\theta)\hat A_t, {\rm clip}(R_t(\theta), 1-\epsilon, 1+\epsilon)\hat A_t\right)\right],
\end{equation} where $R_{t}(\theta) = \pi_\theta(a_t|s_t)/\pi_{\theta_\textit{old}}(a_t|s_t)$ and $\epsilon$ is a hyperparameter. The clipping ensures the gradient steps do not overshoot in the policy parameter space in each policy update.  

\begin{algorithm}[ht!]
    \caption{Policy Optimization with Constraint Advantage Regularization (POCAR)}\label{alg:pocar}
    \begin{algorithmic}[1]
    \State Initialize policy network $\pi_\theta$ and value function network $V_\phi$
    \For{$k = 1, 2 \dots $} \label{state:for1}
        \State Initialize replay buffer $B$ as $\emptyset$
        \For{$\text{episode } = 1, \dots, E$}
            \For{$t = 1, 2 \dots, T$}
                \State \small Sample $a_t \sim \pi_\theta(a_t | s_t)$
                \State \small Sample $s_{t+1} \sim f(s_{t+1} | s_t, a_t)$
                \State \small Compute $\Delta_t \equiv \Delta(s_t) \text{ and } \Delta_{t+1} \equiv  \Delta(s_{t+1})$ \label{state:deltacomputations}
                \State \small $B \gets B \cup \{s_t, a_t, r_t, s_{t+1}, \Delta_t, \Delta_{t+1}\}$ \label{state:updateB}
            \EndFor
        \EndFor
        \For{each policy gradient step}
            \State Sample mini-batch $\{s_t, a_t, r_t, s_{t+1}, \Delta_t, \Delta_{t+1}\}^N$ from $B$
            \State \small Compute advantages $\hat{A}(s_t, a_t)$ (using any advantage approximation method) based on $V_{\phi_k}$
            \State {\small $\hat{A}_\beta (s_t, a_t) \gets \beta_0 \hat{A}(s_t, a_t) + \beta_1         \text{min}(0, -\Delta_t + \omega) + \beta_2 
                \begin{cases}
                \text{min}(0, \Delta_t - \Delta_{t+1}) & \text{if } \Delta_t > \omega \\
                0 & \text{otherwise} \\
                \end{cases}$}
            \State \small $R_t(\theta) \gets \pi_\theta(s_t, a_t) / \pi_{\theta_k}(s_t, a_t)$
            \State \small $J^{\rm CLIP}(\theta, \beta) \gets \hat{\mathbb{E}}\left[ \text{min}\left(R_t(\theta) \hat{A}_\beta(s_t, a_t), \rm clip \left(R_t(\theta), 1 - \epsilon, 1 + \epsilon \right) \hat{A}_\beta (s_t, a_t) \right)\right]$
            \State \small $\theta \gets \theta + \alpha_\theta \nabla_\theta J^{\rm CLIP}(\theta, \beta)$
            \State \small $\phi \gets \phi + \alpha_\phi \nabla_\phi \hat{\mathbb{E}}\left[ (V_\phi(s_t) - G(s_t))^2 \right]$
            \Comment{$G(s_t) \gets \sum^T_{i=0} \gamma^i r_{t+i}$}
        \EndFor
    \EndFor \label{state:endfor1}
    \end{algorithmic}
\end{algorithm}

\paragraph{Fairness Constraint Advantage Regularization.} We modify the PPO algorithm to be fairness-constrained during policy optimization in the following way. First, in addition to the original MDP with only the overall utility as reward function $r$, we declare a separate fairness metric $\Delta:S\rightarrow \mathbb{R}^{\geq}$ as a function of the state, where smaller $\Delta$-value indicate better satisfaction of the fairness constraints. We then task the PPO algorithm to learn a decision policy to maximize the expected discounted future return $J$, while regulating the advantage term to decrease $\Delta$-value in each step. Suppose fairness at a particular time step can be modeled by the fairness constraint metric $\Delta(s_t)$. We minimize this term by modifying the advantage function in PPO to:
\begin{equation} \label{equation::advantage_regularization}
\hat{A}_\beta (s_t, a_t) = \beta_0 \hat{A}(s_t, a_t) + \beta_1 \text{min}(0, -\Delta(s_t) + \omega) + \beta_2 
    \begin{cases}
    \text{min}(0, \Delta(s_t) - \Delta(s_{t+1})) & \text{if } \Delta(s_t) > \omega \\
    0 & \text{otherwise} \\
    \end{cases}
\end{equation}
where $\hat{A}(s_t, a_t)$ is the original PPO advantage. The two terms added to the original advantage estimator are designed as follows. 

\noindent $\bullet$ The value-thresholding term $\min(0, -\Delta(s_t) + \omega)$ penalizes the advantage value for $(s_t,a_t)$ if the fairness metric $\Delta(s_t)$ is too high. That is, if it is higher than some threshold $\omega$, then this term evaluates to a negative value. On the other hand, if $\Delta(s_t)$ is already less than $\omega$, then the term evaluates to zero, and does not affect the advantage value. On a high level, this term transforms the advantage landscape by adding a region of attraction around some equilibrium point, which specifies a tolerable violation of $\Delta$.

\noindent $\bullet$ The decrease-in-violation term $\text{min}(0, \Delta(s_t) - \Delta(s_{t+1}))$ is activated when the value of $\Delta(s_t)$ is larger than the tolerable threshold $\omega$. When that is the case, this term penalizes the overall advantage if the degree of fairness violation does not decrease in the transition from $s_t$ to $s_{t+1}$. If either the violation is decreasing or is less than $\omega$, then this term does not affect the overall advantage. The decrease-in-violation term imposes the soft requirement of negative-definiteness of Lie derivatives in the standard Lyapunov conditions. Here we use the the finite difference of $\Delta(s_t)-\Delta(s_{t+1})$ to measure the dynamics of the fairness metric over time. 

For both of the advantage regularization terms, the hyperparameters $\omega, \beta_0, \beta_1,$ and $\beta_2$ are positive numbers that can be tuned. The design of these terms uses the idea of Lyapunov stability~\cite{lyapunovbook}, which is a powerful framework in control theory for stabilization problems such as reducing the error signal, which can be used to model the violation of the fairness requirement over time.

The full pseudocode for policy optimization with advantage regularization is shown in Algorithm \ref{alg:pocar}. At each iteration of the policy update loop (Lines \ref{state:for1}-\ref{state:endfor1}), we sample trajectories from the system $f$, and compute the fairness constraint $\Delta(s_t)$ and $\Delta(s_{t+1})$ (Line \ref{state:deltacomputations}) to save into buffer $B$ alongside state $s_t$, action $a_t$, reward $s_t$, and next state $s_{t+1}$ (Line \ref{state:updateB}). In each policy update step, we modify the advantage function to include the two additional fairness regularization terms as described above. 

\paragraph{Connection with Lyapunov Methods for Stabilization Control.}

In the proposed methods, we are essentially considering the problem of ensuring long-term fairness as a control problem: the goal is to obtain a control policy to regulate the entire system to improve its satisfaction of the fairness properties over time. Consequently, we are implicitly using control-theoretic methods from the Lyapunov stability framework \cite{lyapunovbook, DBLP:conf/icra/ChangG21, Richards2018TheLN, LIU2020108758, bobiti2018stabilityverification} in our approach, because long-term fairness is essentially about stability properties: we wish to control the system such that the violation of fairness reduces over time, and ideally gets controlled to be under a certain threshold. This connection makes it possible to use Lyapunov methods to derive the regularization terms. A brief introduction to the theory behind Lyapunov stability is provided in Section \ref{appendix:lyapunovintro} of the Appendix.

\section{Experiments}

We evaluate the proposed approaches in three case studies as proposed in~\cite{DBLP:conf/fat/DAmourSABSH20, DBLP:journals/corr/atwood2019treatment}, including attention allocation for incident monitoring, credit approval for lending, and disease control in population networks. In all environments, we compare the performance of policies found by the proposed methods with known human-designed policies, or strategies designed by humans that do not involve any learning-based algorithmic design. We also compare different ways of imposing the fairness constraints in the policy optimization procedures, as well as the Constrained Policy Optimization (CPO)~\cite{achiam2017constrained} approach, to evaluate the effectiveness of advantage regularization techniques. 

First, we broadly define several common agents across the experiments. The greedy baseline agent, as the name suggests, is a human-designed policy that maximizes for some objective without any fairness constraints. The CPO agent is a closely-related RL baseline for us to compare our advantage regularization approach with. For our proposed method, we evaluate three variations of PPO-based policy optimization: Greedy PPO (G-PPO) greedily maximizes the objective without any fairness constraints, Reward-Only Fairness Constrained PPO (R-PPO) is fairness constrained only at the reward level by adding some variation of $-\text{max}(0, \Delta_t - \omega)$ to the reward function where $\omega$ thresholds the fairness constraint, and Advantage Regularization PPO (A-PPO) is fairness constrained only at the policy gradient level by applying Equation \ref{equation::advantage_regularization} to PPO during training. Going forward, we will be referring to these three PPO variations and their definitions by their short-hand notations (G-PPO, R-PPO and A-PPO) for all experiments.

\subsection{Case Study: Attention Allocation for Incident Monitoring}

We first consider the problem of attention allocation in a dynamic setting, formulated as an MDP in~\cite{DBLP:conf/fat/DAmourSABSH20}. In this problem, an agent is tasked with discovering incidents across several sites, and does not have a large enough attention span to cover all incidents occurring at any given moment. The site incident rates increase or decrease proportional to the allocated site attention, making the environment dynamic. We find that training an RL agent in this environment outperforms the human-designed policy in both utility and fairness over time. Moreover, we modified the environment to be harder, where the previously known policies perform significantly worse. We show that policy optimization with advantage regularization continues to work well in the harder environments as well. 

\paragraph{Environment.} Let $a_{kt}$ be the discrete attention allocated toward site $k$ at time $t$, and $R_{kt}$ be the incident rate. At each time step, the agent assigns $N$ discrete units of attention across $K$ sites. The number of incidents occurred at each site is sampled $y_{kt} \sim \mbox{Poisson}(R_{kt})$ and number of incidents discovered at each site is $\hat{y}_{kt} := \min(a_{kt}, y_{kt})$. The incident rates change proportional to the attention allocated to each site: if $a_{k,t} = 0$, then $R_{k, t+1} = R_{k,t} + d$, otherwise $R_{k, t+1} = R_{k,t} - d\cdot a_{k,t}$, where $d$ is the parameter that controls how dynamic the environment is. The reward function promotes incident discovery and penalizes for incidents missed. It is defined
$r(s_t) = \zeta_0 \sum_k \hat{y}_{kt} - \zeta_1 \sum_k (y_{kt} - \hat{y}_{kt}),
\label{equation::attention_allocation_objective}$
where $\zeta$ is a vector of reward weights. 

\paragraph{Fairness Constraint.} The fairness metric to be minimized is:

\begin{equation}
    \Delta(s_t) := \text{max}_{k,k'} \bigg | \frac{\sum_t \hat{y}_{kt}}{\sum_t y_{kt} + 1} - \frac{\sum_t \hat{y}_{k't}}{\sum_t y_{k't} + 1} \bigg |.
\end{equation}

which measures the maximum difference in the ratios between total incidents discovered and total incidents occurred across all sites.

\paragraph{Agents.}
Our baseline agents include the CPO agent and the purely greedy agent~\cite{DBLP:conf/fat/DAmourSABSH20}, which discovers the most incidents and is considered the most fair out of all agents as evaluated in~\cite{DBLP:conf/fat/DAmourSABSH20}. The purely greedy agent approximates each site's incident rate and sequentially allocates each unit of attention to the site that has the highest probability of inciting an incident to occur. We compare these baselines with the G-PPO, R-PPO, and A-PPO variations defined above.

\paragraph{Results.}

\begin{figure}[h!]
    \centering
    \includegraphics[width=1\textwidth]{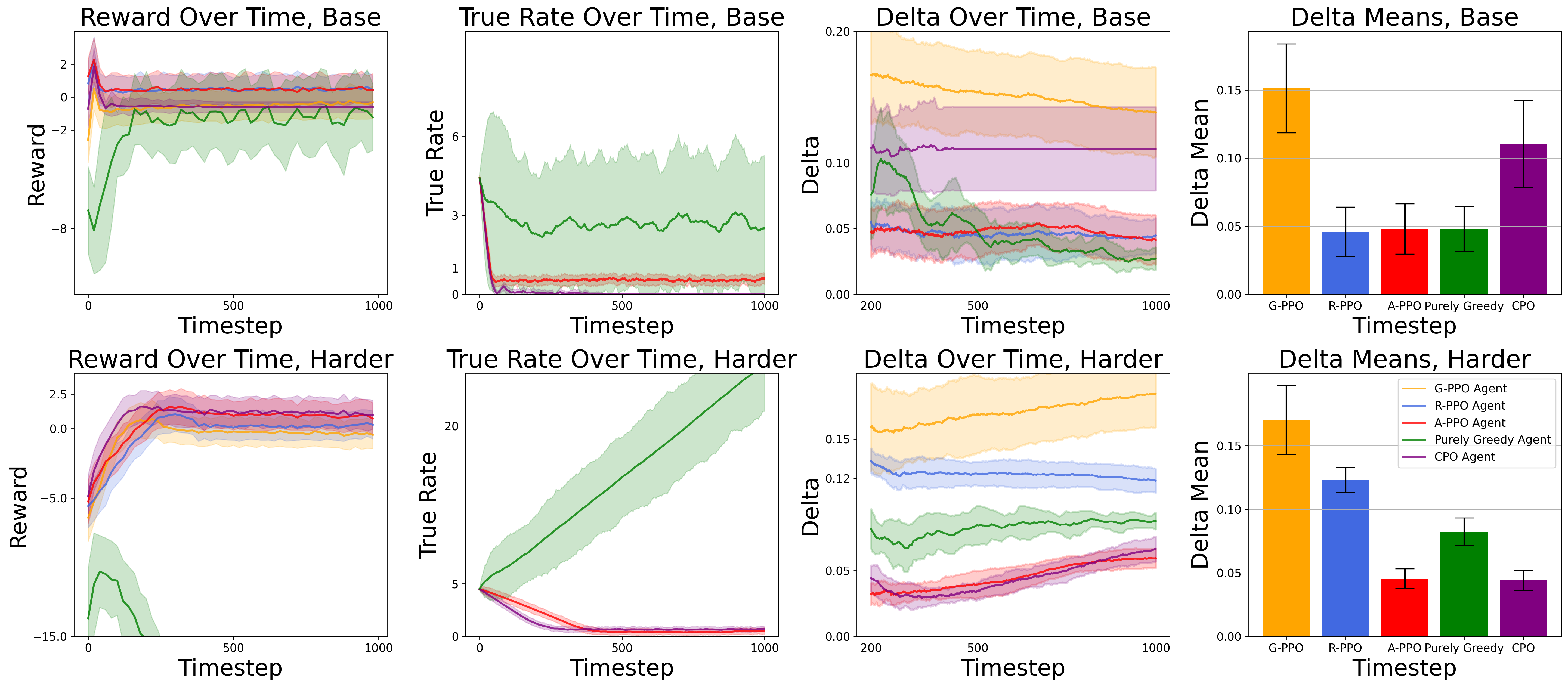}
    \caption{The agent reward over time, average incident rate across all sites over time, and delta summary evaluated over 10 trials for the base and harder attention allocation environments. All policy optimization algorithms, including the CPO agent, outperform the purely greedy agent in both reward and fairness. In the harder environment, the purely greedy policy collapses in performance, and we observe how advantage regularization can be a robust form of imposing fairness.}
  \label{fig::attention_allocation::reward_and_true_rate_over_time_and_delta_summary}
\end{figure}

We begin our experiments with first evaluating on the base attention allocation environment, where $d = 0.1$, $N = 6$, and $k = 5$. We observe more interesting results by evaluating on the harder attention allocation environment, i.e. changing $k = 10$.

In the base attention allocation environment in Figure \ref{fig::attention_allocation::reward_and_true_rate_over_time_and_delta_summary}, every PPO variation outperforms the baseline purely greedy agent in terms of reward over time. G-PPO does not perform as well as R-PPO or A-PPO, which is expected because the evaluation metric includes the fairness constraint term. Although A-PPO also does not have a fairness term in the objective, having advantage regularization in this setting allows A-PPO to perform just as well as R-PPO. In terms of fairness, we observe that G-PPO is the least fair agent as expected. Although the purely greedy agent does become the most fair agent at the end of the trajectory, R-PPO and A-PPO are more consistent in how fair they are on average over the entire trajectory. We observe that the CPO agent does not perform well on the base environment, because its incident rates get pushed to 0 early on in the trajectory in Figure \ref{fig::attention_allocation::reward_and_true_rate_over_time_and_delta_summary}.

To analyze why PPO outperforms the purely greedy agent in reward and fairness, we need to interpret their behaviors in Figure \ref{fig::attention_allocation::reward_and_true_rate_over_time_and_delta_summary}. Here, we observe that the average incident rate for the purely greedy agent is much higher than that of the PPO agent, so it will miss more incidents. The purely greedy agent's incident rate also has the highest standard deviation, which is consistent with the greedy agent's behavior of focusing all its attention on each site sequentially. By reducing the incident rate of a single site at a time but increasing the incident rates at all other sites, the purely greedy agent pushes the incident rates across sites to oscillate. After becoming well-trained, the PPO agents find that refining control on the incident rates across sites in this range allow the agents to maximize on the objective while being consistently fair.

Since any attention placed on a site means the site's incident rate goes down, it is possible to push all site incident rates to 0 if $K \geq N$. However, what happens if we modify the environment such that $N < K$ and the agent is unable to decrease the incident rates across all sites at once? After evaluating on this harder environment, our results become more interesting.

In the second row of Figure \ref{fig::attention_allocation::reward_and_true_rate_over_time_and_delta_summary} A-PPO now performs significantly better than R-PPO, both of which achieve higher rewards over time than G-PPO. The CPO agent also does very well, and actually exceeds A-PPO in reward over time in the beginning of the trajectory. However, the purely greedy policy collapses in performance. To understand the cause of the baseline failing, we look at Figure \ref{fig::attention_allocation::reward_and_true_rate_over_time_and_delta_summary}. We see that the average incident rate in the purely greedy agent's evaluation shoots up uncontrollably. This is a stark contrast from its evaluation on the base environment, and reveals how this policy may be sensitive to slight changes in the problem setup. The PPO agents remain relatively consistent in their strategy, and still push the incident rates to a range in which they can exert a more refined control on the sites. In terms of fairness, incorporating advantage regularization allows A-PPO to be more fair than R-PPO as observed in Figure \ref{fig::attention_allocation::reward_and_true_rate_over_time_and_delta_summary}. Purely greedy agent, despite collapsing in performance, is more fair than the R-PPO and G-PPO. The CPO agent is also very fair, and actually matches performance with A-PPO. This tells us that our A-PPO is at least as good as the CPO agent, and can be a very favorable alternative considering how theoretically it takes longer to train the CPO agent. In fact, in this experiment we found that empirically our PPO agent trained 2-3x faster on average than the CPO agent.

We find that in this experiment, decision policies obtained from policy optimization methods clearly outperform the human-designed policies in both reward and fairness. Incorporating advantage regularization for fairness performs better than only having a fairness term in the reward. Advantage regularization also appears to be more robust since in a harder environment, A-PPO is able to perform the best.

\subsection{Case Study: Credit Approval for Lending}

Next, we consider credit approval for lending in a dynamic setting~\cite{DBLP:conf/fat/DAmourSABSH20}. In this environment, the agent plays the role of the bank and is tasked with deciding whether to accept or reject loan requests from a pool of applicants. These decisions affect the underlying population distribution for future time steps. In our experiments, we find that PPO agents are able to perform well without the oracle access to the underlying environment, with performance on par with the baseline strategies that need to make use of such ground-truth information. A-PPO is still the best policy optimization approach, though the gap between A-PPO and R-PPO is smaller than the previous example because the task is simpler. 

\paragraph{Environment.}

The agent is presented a stream of loan applicants, each with a discrete credit score $C \in \{1, 2, ..., C_{\text{max}}\}$ and some group membership variable $g \in \{1, 2\}$. Group 2 is considered disadvantaged   compared to Group 1 with a lower mean of the initial credit score compared to Group 2. Applicants are uniformly sampled from these two groups. The environment dynamics change in response to the bank decision (accept or reject) and applicant decision (repay or default). If a loan request is accepted, the probability that an applicant pays the bank back is given by the underlying deterministic function of credit score $\eta(C)$. The higher the applicant credit score, the more likely they are to pay the bank back. The reward function encourages high bank profits and is defined $r(s_t) = \zeta_0 (B_{t+1} - B_t)$ where $B$ represents the bank's total cash reserve.

\paragraph{Fairness Constraint.}

The fairness constraint is defined as $\Delta(s_t) = \text{max}_{g, g'} |\text{TPR}_{gt} - \text{TPR}_{g't}|$, where $\text{TPR}_g$ is the true positive rate $\frac{\text{TP}}{\text{TP} + \text{FN}}$ for group $g$. TP is the number of true positives corresponding to when the agent accepts a loan request and the loan is repaid, and FN is the number of false negatives corresponding to when the agent rejects a loan request and the applicant would have repaid.

\paragraph{Agents.}
We have two baseline policies following the human-designed lending policies in \cite{DBLP:conf/fat/DAmourSABSH20}. Our first baseline agent is the greedy policy that attempts to maximize bank profits. The second baseline agent is the equality of opportunity (EO) agent \cite{NIPS2016_9d268236}, which maximizes bank profits with the constraint of equalizing the TPRs between the two groups. We note that to simplify the task, \cite{DBLP:conf/fat/DAmourSABSH20} gives the EO agent oracle access to the underlying distribution of credit scores amongst both groups and the exact repayment probabilities function $\eta$. We later show that RL is able to achieve similar, if not better, performance compared with the EO agent even without access to the underlying environment information. We also include the CPO agent as our secondary fairness-constrained deep RL baseline.

The PPO agent observation is the applicant data consisting of their credit score $C$ and group membership variable $g$. The action is a binary decision for whether to accept or reject an applicant's loan request. Like in the attention allocation environment, we evaluate the three previously defined PPO variations (G-PPO, R-PPO, A-PPO), but for added stability during training, we apply a min-max normalization to each term in Equation \ref{equation::advantage_regularization} for A-PPO before performing their weighted sum.

\paragraph{Results.}

\begin{figure}[h!]
    \centering
    \includegraphics[width=1\textwidth]{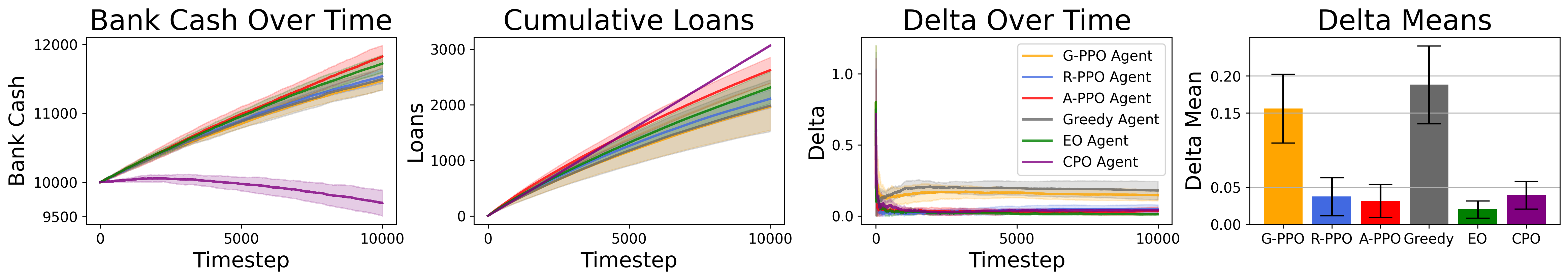}
    \caption{The bank cash over time, loans over time, and delta plots from evaluating on the lending experiment for 10 trials. We note that in this experiment, the PPO variations make more profits on average than the baseline algorithms. Although the EO policy is the most fair and makes a big profit over time, the A-PPO still out-earns the EO policy and is able to maintain a competitive level of fairness without having oracle access to underlying environment information. The CPO agent is one of the most fair agents, but fails to profit.}
  \label{fig::lending::lending_bank_cash_and_loans_over_time_and_delta_summary}
\end{figure}

We run experiments with $C_{\text{max}} = 7$. From Figure \ref{fig::lending::lending_bank_cash_and_loans_over_time_and_delta_summary}, it is interesting to see that all the fairness constrained agents except CPO make a bigger profit than the fairness unconstrained agents like G-PPO and greedy. To understand this trend, we look at the average number of loans made across groups by each agent in Figure \ref{fig::lending::lending_bank_cash_and_loans_over_time_and_delta_summary}. We see that the profit an agent makes roughly correlates to how many loans they give out on average. This makes sense since an agent cannot make money unless it gives out loans, and the agents that give out more loans can be interpret as more willing to take risks for higher rewards. 

\subsection{Case Study: Infectious Disease Control in Population Networks}

In our third experiment, we analyze the precision disease control problem \cite{DBLP:journals/corr/atwood2019treatment}. In this problem, the agent's task is to vaccinate individuals within a social network to minimize the spread of a disease. In this experiment, we observe that the PPO variations still outperform the human-designed policies in both utility and overall fairness. Integrating advantage regularization appears to have an equal effect as constraining fairness at the reward level only, which we analyze later. The CPO agent fails to perform well on this environment.

\paragraph{Environment.}

A social network $N$ is composed of individuals $V$ connected with edges $E$. Each individual has one of three health states defined by $\{S, I, R\}$ for susceptible, infected, and recovered. The health of the network is then characterized by vector $H \in \{S, I ,R\}^{|V|}$. At the beginning of a trajectory, a random individual in $N$ gets infected. For each time step, the infection spreads from an infected individual to a susceptible individual $v$ under $p^v_{S \rightarrow I} = 1 - (1 - \tau)^{\#I(v, N)}$, where $\tau \in [0, 1]$  is the probability of spreading and $\#I(v, N)$ is the number of infected neighbors of susceptible individual $v$ in $N$. Then, probability of recovery is given by $p_{I \rightarrow R} = \rho$. At each time step, the agent is able to allocate one vaccine to give to an individual. Any individual can receive an unrestricted number of vaccinations, but vaccinating an individual will only have an effect if they are in the susceptible state. Vaccinations do not roll over over time steps, but the agent is able to choose not to vaccinate anyone at a time step. The reward function promotes the healthiness of a population and is given $r(s_t) = \zeta_0 \cdot (\sum_{i=1}^{|V|}\mathds{1}(H_{it} \neq I))/|V|$.

We define the notion of a community in $N$ by applying the Girvan-Newman algorithm \cite{girvannewman} on the graph. This community detection algorithm progressively removes edges from the graph by betweenness, which is the number of shortest paths between two nodes that travel through an edge. We apply this algorithm once to obtain two communities (visualized in the Appendix).

\paragraph{Fairness Constraint.}
The fairness constraint is defined as
\begin{equation}
    \Delta(s_t) = \text{max}_{c,c'} \bigg | \frac{\sum_t \text{vaccinations given}_{ct}}{\sum_t \text{newly infected}_{ct} + 1} - \frac{\sum_t \text{vaccinations given}_{c't}}{\sum_t \text{newly infected}_{c't} + 1} \bigg |,
\end{equation}
where $c$ is a community. This fairness constraint is inspired by that from the attention allocation problem, and focuses on minimizing the differences in the ratio of vaccinations to newly infected individuals over time across communities. 

\paragraph{Agents.} The PPO agent observation is the health states $H$ of the network. The action is a number $\{1, 2, ..., |V|\} \cup \{\varnothing\}$ where $\varnothing$ corresponds to not vaccinating anyone. We again use the three previously defined PPO agents (G-PPO, R-PPO, A-PPO). Similar to the previous case study, a min-max normalization is applied to each term in Equation \ref{equation::advantage_regularization} for A-PPO. We define two variations of the human-designed policies in~\cite{DBLP:conf/fat/DAmourSABSH20}. First, the random baseline will randomly select a susceptible individual to vaccinate, and if there are no more susceptible individuals, then it will try to vaccinate someone randomly in the graph regardless of status. We also define the max neighbor (denoted as "Max") baseline that will vaccinate susceptible individuals sorted by number of neighbors in the graph, and do the same for non-susceptible individuals if there are no more susceptible individuals in the graph.

\paragraph{Results.}

\begin{figure}[h!]
    \centering
    \includegraphics[width=1\textwidth]{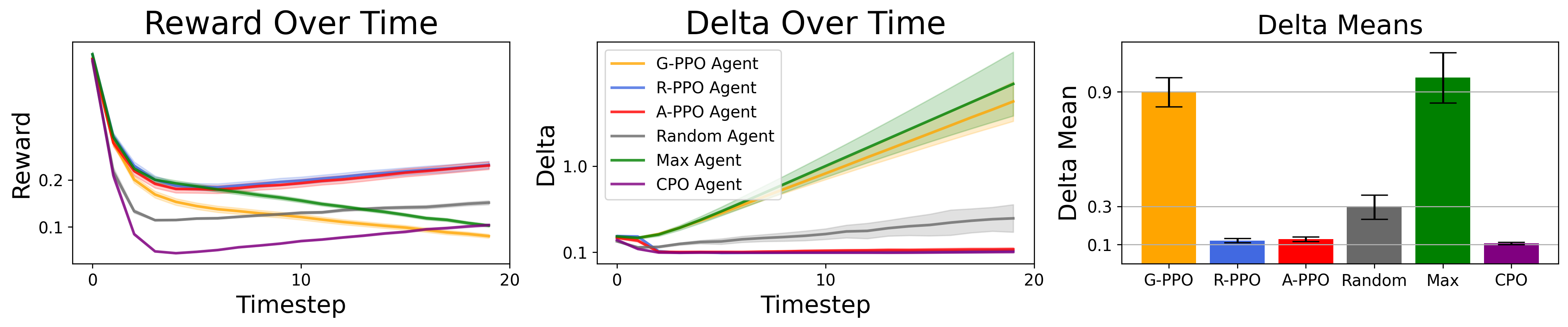}
    \caption{The reward over time and delta plots generated by evaluating on the precision disease control environment for 200 trials. We observe that all PPO policies outperform the baseline policies, with both A-PPO and R-PPO being equally the most fair.}
  \label{fig::infectious::reward_over_time_and_delta_summary}
\end{figure}

We run experiments on the Karate Club graph for 20 time steps per trajectory with $\tau = 0.5$ and $\rho = 0.005$. We set the initially infected individual to be the most connected node in the graph for the most interesting results, and impose a burn-in period where the infection is allowed to freely spread for one time step.

As seen in Figure \ref{fig::infectious::reward_over_time_and_delta_summary}, the PPO agents again outperform the baseline agents in reward and fairness. A-PPO and R-PPO both collect an equal amount of rewards, with R-PPO being slightly more favorable. The G-PPO and max agents' rewards fall off as they fail to satisfy the fairness evaluation objective over time. We observe that the random agent begins with a lower reward, but increases over time, which can be explained by the fairness analysis next. The CPO agent is observed to actually perform the worst in terms of utility. In Figure \ref{fig::infectious::reward_over_time_and_delta_summary}, A-PPO and R-PPO are also the most fair, as they both quickly drive the fairness constraint toward zero over time. The random agent is the next fairest agent, which can be explained by the fact that the random agent does not discriminate between groups when deciding who to vaccinate next. The G-PPO and max agents actually increase linearly in delta over time, which also makes sense since neither agent is constrained for fairness. Although CPO did not perform well in reward, it is actually one of the most fair agents evaluated here.

This experiment provides an additional insight into how RL agents can outperform human-designed policies. Here, we observe that A-PPO and R-PPO outperform all other agents in terms of reward and fairness. Although advantage regularization appears to not have any significant improvement over fairness constraining at the reward level, this experiment still serves as an example for how our approach can perform at least as well as reward-only fairness constraining, and how it can be more robust of a procedure than CPO.

\section{Conclusion}

We proposed new methods to design fairness-constrained decision-based neural control policies in long-term fairness problems. We augmented our approach with advantage regularization by formulating fairness as a stability property within policy optimization. We demonstrated the benefits of these methods in several dynamic environments for long-term fairness. Future work includes extending this framework to other policy optimization algorithms and investigating how our approach can be extended with stricter fairness-constrained methods. 

We remark that our approach only guarantees optimality under the definitions of the objective and the fairness constraints, and the optimal way to minimize this constraint does not necessarily correspond to the most socially acceptable way to solve long-term fairness problems. Thus, fairness obtained through our proposed methods is limited to the design formulation of the long-term fairness problem. Consequently, it is vital to correctly formulate the objectives and fairness requirements while being aware of how the fairness requirements imposed on the policy may induce unexpected and potentially unfair behaviors from a different perspective.

\section{Acknowledgement}
This material is based on work supported by DARPA Contract No. FA8750-18-C-0092, AFOSR YIP FA9550-19-1-0041, NSF Career CCF 2047034, NSF CCF DASS 2217723, and Amazon Research Award. We appreciate the valuable feedback from Ya-Chien Chang, Milan Ganai, Chiaki Hirayama, Chenning Yu, Hongzhan Yu, Yaoguang Zhai, Ruipeng Zhang, and the anonymous reviewers. 

\bibliographystyle{plain}
\bibliography{ref}

\newpage
\appendix

\begin{appendices}


\section{An Introduction to Lyapunov Stability}
\label{appendix:lyapunovintro}

We first define an $n$-dimensional dynamical system $\Phi$ by:

$$\dot{x}(t) = f(x(t), u(t)), u(t) = g(x(t))$$

where $x(t) \in D$ is a state vector at time $t \in \mathbb{R}$ in the state space $D \subseteq \mathbb{R}^n$, $g : D \rightarrow \mathbb{R}^m$ is a control function, and $f : D \rightarrow \mathbb{R}^n$ is a Lipschitz-continuous vector field.

Suppose system $\Phi$ has an equilibrium point at $x_e$. The system is stable at $x_e$ if for all $\epsilon\in\mathbb{R}^+$, there exists some $\delta(\epsilon)\in\mathbb{R}^+$ such that $||x(t) - x_e|| < \epsilon$ for all $t \ge 0$ if $||x(0) - x_e|| < \delta$. In other words, if an initial point $x(0)$ starts at some distance $\delta$ from equilibrium point $x_e$, any point along all possible solution trajectories from $x(0)$ to $x_e$ should be smaller than some distance $\epsilon$ from $x_e$. To take this stability notion one step further, we say system $\Phi$ is locally asymptotically stable around $x_e$ if $\lim_{t\rightarrow\infty} x(t) = 0$ for all $||x(0) - x_e|| < \delta$. 

Next, we explore the definition of the Lyapunov function and Lie derivative. Using the existing system setup, let $V : D \rightarrow \mathbb{R}$ be a continuously differentiable function. $V$ is a Lyapunov function if $V(0) = 0$ and $L_fV(x(t)) < 0$ and $\forall x \in D \setminus \{0\}, V(x(t)) > 0$. The Lie derivative of $V$ over $f$ is defined:

$$L_fV(x(t)) = \sum^n_{i=1} \frac{\partial V}{\partial t} = \sum^n_{i=1} \frac{\partial V}{\partial x_i} \frac{dx_i}{dt} = \sum^n_{i=1} \frac{\partial V}{\partial x_i} \dot{x}(t)$$

At a high level, the Lyapunov function $V$ defines a field of attraction around some equilibrium point, and the Lie derivative $L_fV(x(t))$ measures the rate of convergence of V over time along the system dynamics of $x(t)$ to its equilibrium point $x_e$. If this Lyapunov function can be defined, system $f$ is asymptotically stable at $x_e$. 

One issue with formulating the Lie derivative in the context of Reinforcement Learning (RL) training is that computing it requires full access to system dynamics $f$, which the RL policy in training does not have access to. Thus, we must approximate the Lie derivative along sampled trajectories of the dynamical system during training:

$$L_{f, \Delta t} V(x(t)) = \frac{1}{\Delta t} (V(x(t+1)) - V(x(t)))$$
where $$\lim_{\Delta t \rightarrow \infty} L_{f, \Delta t} V(x(t)) = L_{f}V(x(t))$$

\section{Hyperparameters for Case Studies}

We used the following hyperparameters in the training procedures. Table~\ref{tab:attention_ppo_table}, Table~\ref{tab:lending_ppo_table} and Table~\ref{tab:disease_ppo_table} show the hyperparameters used for PPO variations on attention allocation, lending, and infectious disease control environments, respectively. Note that for the lending and precision disease control environments, our hyperparameters are chosen with respect to a min-max normalization applied to each advantage term. 

\begin{table}[h!]
\centering
 \begin{tabular}{|c | c c c c c c|} 
 \hline
    PPO Agent & $\zeta_0$ & $\zeta_1$ & $\zeta_2$ & $\beta_0$ & $\beta_1$ & $\beta_2$ \\ [0.5ex] 
 \hline
 Greedy (G-PPO) & 1 & 0.25 & 0 & 0 & 0 & 0 \\ 
 Reward-Only Fairness Constrained (R-PPO) & 1 & 0.25 & 10 & 0 & 0 & 0 \\
 Advantage Regularized (A-PPO) & 1 & 0.25 & 0 & 0.05 & 0.32 & 0.63 \\ [1ex] 
 \hline
 \end{tabular}
 \vspace*{3mm}
 \caption{The hyperparameters used for each PPO variation during training on the base and harder attention allocation environments.}
 \label{tab:attention_ppo_table}
\end{table}

\begin{table}[h!]
\centering
 \begin{tabular}{|c | c c c c c|} 
 \hline
    PPO Agent & $\zeta_0$ & $\zeta_1$ & $\beta_0$ & $\beta_1$ & $\beta_2$ \\ [0.5ex] 
 \hline
 Greedy (G-PPO) & 1 & 0 & 0 & 0 & 0 \\ 
 Reward-Only Fairness Constrained (R-PPO) & 1 & 2 & 0 & 0 & 0 \\
 Advantage Regularized (A-PPO) & 1 & 0 & 1 & 0.5 & 0.5 \\ [1ex] 
 \hline
 \end{tabular}
 \vspace*{3mm}
 \caption{The hyperparameters used for each PPO variation during training on the lending environment.}
 \label{tab:lending_ppo_table}
\end{table}

\begin{table}[h!]
\centering
 \begin{tabular}{|c | c c c c c|} 
 \hline
    PPO Agent & $\zeta_0$ & $\zeta_1$ & $\beta_0$ & $\beta_1$ & $\beta_2$ \\ [0.5ex] 
 \hline
 No Fairness Constraints (N PPO) & 1 & 0 & 0 & 0 & 0 \\ 
 Only Reward Fairness Constraint (R PPO) & 1 & 0.1 & 0 & 0 & 0 \\
 Only Advantage Fairness Constraint (A PPO) & 1 & 0 & 0.6 & 0.15 & 0.25 \\ [1ex] 
 \hline
 \end{tabular}
 \vspace*{3mm}
 \caption{The hyperparameters used for each PPO variation during training on the precision disease control environment.}
 \label{tab:disease_ppo_table}
\end{table}

\section{Social network for Infectious Disease Case Study}

\begin{figure}[h!]
    \centering
    \includegraphics[width=0.6\textwidth]{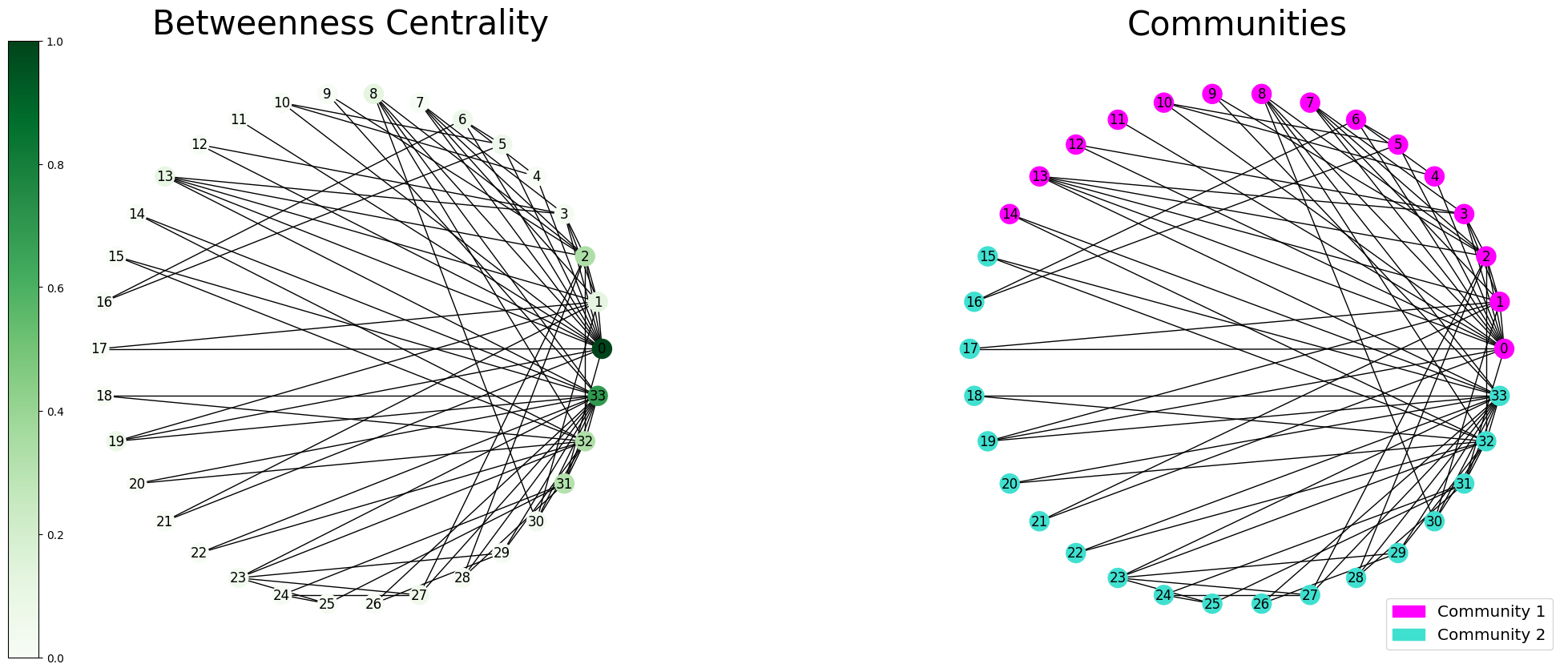}
    \caption{[Left] Node betweenness centrality visualized for the Karate Club graph in the precision disease control environment, where the intensity of the node positively correlates with its betweenness centrality value. [Right] Communities for the Karate Club graph in the precision disease control environment. These communities are obtained by applying the Girvan-Newman community detection algorithm once on the graph.}
  \label{fig:infectious_betweenness_and_communities}
\end{figure}

We obtain our notion of a community in the Karate Club graph using the Girvan-Newman community detection algorithm. We define edge betweenness as the number of shortest paths between two nodes that travel through an edge. In this algorithm, each edge is computed for its edge betweenness. Then, the edge with the highest betweenness is removed to reveal two communities seen on the right in Figure~\ref{fig:infectious_betweenness_and_communities}. Node betweenness centrality is defined as the combined fraction of all shortest paths between pairs that pass through a node, and can be visualized on the left in Figure~\ref{fig:infectious_betweenness_and_communities}. Although betweenness centrality is distinct from edge betweenness and is not a part of the Girvan-Newman algorithm, we include it to provide more insight into the underlying structure of the Karate Club Graph.

\end{appendices}

\end{document}